\documentclass[a4paper]{article} 

\usepackage{setspace}

\usepackage[english]{babel}
\usepackage{graphicx} 

\usepackage{cite}

\usepackage{latexsym} 
\usepackage{amsfonts} 
\usepackage{amsmath}

\usepackage{multirow}

\usepackage{enumitem}

\usepackage[utf8]{inputenc}
\usepackage[T1,T2A]{fontenc}

\usepackage{layout}

\usepackage[left]{lineno}

\usepackage{color, soul} 
\setulcolor{yellow}

\usepackage[colorlinks=true, citecolor=blue]{hyperref} 

\usepackage[nameinlink]{cleveref}

\title{Automatic measurement of coverage area of  water-based pesticides-surfactant formulation   on plant leaves using deep learning tools. }

\author{Fabio Grazioso$^{*}$, Anzhelika A. Atsapina,\\  Gardoon L. O. Оbaeed, Natalia A. Ivanova$^{*}$\\
	Photonics and Microfluidics Lab, Tyumen State University, Volodarskogo 6,\\ Tyumen 625003, Russia\\
* corresponding authors: [f.grazioso, n.a.ivanova]@utmn.ru}

\begin{document}

\maketitle

\date{\vspace{-5ex}}

\begin{@twocolumnfalse}
	\begin{abstract}
		A method to efficiently and quantitatively study  the delivery of a pesticide-surfactant formulation in water solution over plants' leaves is presented. Instead of measuring the contact angle, the surface of the leaves' wet area  is used as key parameter. To this goal, a Deep Learning model has been trained and tested, to automatically measure the surface of area wet with water solution over cucumber leaves, processing the frames of video footage.
		We have individuated an existing Deep Learning model, reported in literature for other applications, and we have  applied it to this different task. We present  the measurement technique, some details of the Deep Learning model,  its training procedure  and its image segmentation performance. Finally, we report the results of the wet area's surface measurement as a function of the concentration of a surfactant in the pesticide solution.
	\end{abstract}

agriculture.

\end{@twocolumnfalse}

\tableofcontents
\section{Introduction}

The present work is part of the  scientific effort to improve agriculture, using the latest technologies to contribute to what is commonly known as \emph{precision agriculture} \cite{Cisternas2020,daSilveira2021,Wang2022,Rai2023}. One of the goals in this field of research is the study and improvement of the \emph{delivery of pesticides}, which are usually deployed as part of water-based solutions \cite{Wang2019}.

The use of pesticides improves the efficiency and productivity of agriculture  and forestry through  crop protection and  growth stimulation. Surfactants (also referred as adjuvants) are usually added, to enhance the biological activity of pesticide formulations, which allow overcoming the leaf protection barriers, which plants produce in the form of hydrophobic cuticular wax or trichomes  and increase the effective coverage area of pesticide formulations. Surfactants reduce the surface tension of water-based pesticide formulations allowing the adhesion of the spray to the leaf surfaces,  significantly increasing the wetting of foliar surface and enhancing pesticides uptake into leaves \cite{Stevens1993,Taylor2011,Zhang2017,Castro2013,Wang2022a,Jibrin2021,Liu2004,Song2022,Lin2016,Xu2010,Pierce2008,Wang2018a,Fine2017}. 

However, the existing formulations and dosages for the agricultural applications of surfactants do not always give the expected results for the various types of crop leaves. The latter fact stimulates excessive use of both surfactants and pesticides, which on  one hand does not result in a uniform residual coverage of plant leaves after spreading and evaporation of the spray droplets \cite{Ciarlo2012,Xu2011}, and on the other hand leads to an increase in negative impact on the ecosystem, in the form of contamination of soil and water reservoirs, death of  pollinator insects and honeybees \cite{Yu2009,Yu2009a}. In this context, the development of effective surfactant formulations for crop protection, requires  laboratory  and  field measurements of the \emph{wetting area} in order to determine the \emph{optimal ratio} between the coverage area and the surfactant (adjuvants) concentration.
The usual approach found in literature is the use of goniometry to measure the evolution of the contact angle and the diameter of the droplet contact with the leaf surface to get quantitative data about surfactants performance \cite{Song2022}. However, this method does not provide reliable information due to the non-uniform front of the droplet spreading on the surface of leaves with complex morphology. For example, on leaves with longitudinal veining common to cereal crops, the droplet will spread predominantly along the veins, while in case of the reticulate vein pattern (cucurbits crops) the liquid will drain in the vein valleys. Moreover, the presence of trichomes does not allow the droplet contact angle to be clearly defined. Finally, the maximum coverage area of the leaves depends even on which part of the leaf surface the droplet is deposited \cite{Xu2010}.


In light of these considerations, to determine the criterion of surfactant efficiency another approach seems to be preferable: the study of the \emph{spreading area}. For this study, the process of droplet spreading over the leaf surface is recorded from above using a camera and then images (frames) from the video sequence are processed to measure the coverage area. 

Precise measurement of the coverage area of plant leaves as function of wetting and evaporation time after spray deposition and as function of the concentration of chemicals and leaf morphology are strongly required to develop precise rates for the usage of plant protection products and adjuvants. 

Along this line of research on the phenomenon of spreading of aqueous solutions on different types of surfaces, it can be mentioned the following literature from thesame authors of the present work \cite{Ivanova2011,Ivanova2012,Ivanova2017,Ivanova2021,Grazioso-22}. \\


In order to process images  for the measurement of the wet area, it is necessary to perform the \emph{segmentation}: the selection of the parts of the images (pixels) which represent the wet area, differentiating them from all the other parts of the image. Once this segmentation is done, the wet area's surface will be measured counting the total number of pixels representing the wet area. 

One possible approach to perform the segmentation is the manual process: a person manually selects the wet area of each image. This method is probably the most accurate, but it is also the slowest. To put the time required in context, it should be considered that it is easy to record thousands of images, extracting the frames from videos that can last several minutes, in case we want to achieve a high time resolution. So  is the manual processing is unpractical for the processing of such high number of frames.


In order to perform the surface measurement in a high number of images (in the range of thousands), an automatic segmentation method is necessary.

An option for this automation are \emph{numerical comparative algorithms} for the automatic image segmentation, which extract the numerical values of intensity, and/or saturation, luminosity, hue etc. of each pixel and compare them.

Several examples of algorithmic image processing can be found in literature. A simple implementation of numerical algorithm is demonstrated in \cite{Zhu2011}, where the authors evaluate the effectiveness of pesticides formulations by spraying them on a water-sensitive paper and then scanning it. The images obtained are converted into black-and-white images using the ImageJ software and the number of pixels with an intensity corresponding to the wet area is counted. It should be noted that testing formulations on paper does not give adequate information about the behaviour of formulations with surfactants interacting with the surface of plant leaves. 

In a number of studies the image analysis is carried out using the polygonal hand-trace feature found in commercial software  \cite{Lin2016,Xu2010,Pierce2008,Wang2018a,Fine2017}.

In the work reported in \cite{Li2019a}, the algorithm for measuring the maximum leaf coverage area is based on a multi-stage complex image processing procedure and implemented in MatLab. In the wetting experiments, a blue pigment is added to the water-based pesticide formulations. The image processing consists of the extraction from the RGB image of the blue component (related to the blue pigment used), followed by the filtration of the image, the generation of the binary image using a segmentation method, counting the number of pixels by applying a disk mask together with procedure for the noise elimination.

An alternative to numerical algorithms for automatic segmentation is the use of Deep Learning (DL), a specific implementation of \emph{Artificial Intelligence} based on the technology of \emph{Neural Networks}, and the present work is based on this choice. Later in \cref{sec:algo-discuss} it will be discussed a comparison between the algorithmic and DL approaches, and there it will be shown the \emph{drawbacks} of the first one, so that this choice will be motivated.

\section{Methods and measurements}

\subsection{Materials}

For the measurements,  cucumber leaves have been used. The fluid used for the wetting experiments has been a solution of distilled water, with 5\% concentration in volume of a colloidal suspension of silver (the suspension itself has a concentration of 3 g/l). The colloidal silver suspension is a commercial product acting as fungicide and bactericide, purchased from the company AgroKimProm (Russian Federation), and commercialized under the name `Zeroxxe'. Then, an organic, silicon-based super-wetting agent, in particular the commercial product `Majestik', also purchased from  AgroKimProm (Russian Federation), has been added to this solution, in different concentrations.

\subsection{CMC measurement}
\label{sec:CMCmeas}

\begin{figure}
	\centering
	\includegraphics[width=0.8\linewidth]{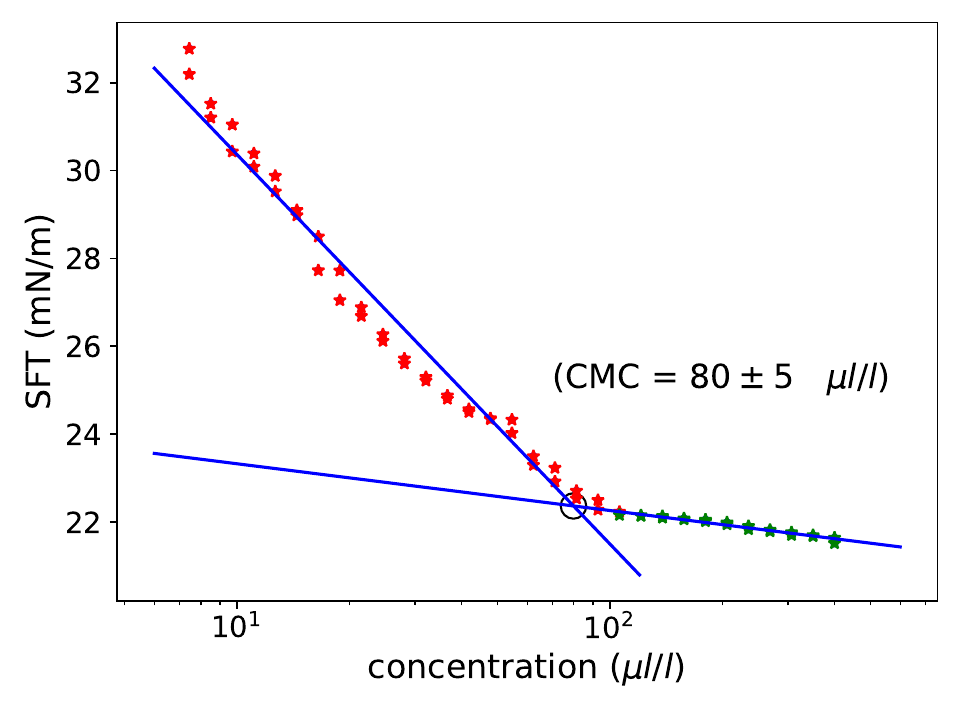}
	\caption{Plot of the data  for  CMC measurement. On the  vertical axis  are reported the values for the surface tension (SFT), which has been measured with the  Wilhelmy plate method, and on the horizontal axis are reported the values for the volume concentration. According to the theory of CMC, the surface tension is supposed to  follow an exponential decay as the concentration of surfactant grows,  until it reaches the critical micelle concentration (CMC), ant then it is supposed to continue linearly. So, the data have been plotted in a semi-logarithmic plane, in order for the exponential part to appear linear. In the figure the datapoints are represented in two different colors to higlight those belonging to the two different regimes (exponential in red, and linear in green). Then, two  linear best fits are performed, for the initial and the final regime (see blue lines), and the CMC is estimated as the concentration at the interception of the two lines (see black circle in the plot). The estimate for the CMC is $ 80 \pm 5 \mu l / l$. It is expressed in $\mu l$ of surfactant, as it is delivered in the vendor's container, per liters of water.}
	\label{fig:CMC}
\end{figure}

In order to have precise information on the surfactant used in the form, a direct measurement of the critical micelle concentration (CMC) of the surfactant has been performed, using the tensiometer model \emph{DCAT 15} from the company DataPhysics Instruments (Germany), based on the Wilhelmy plate method.
In  \cref{fig:CMC} we report a plot of the CMC measurement datapoints, in a semi-logarithmic plot, with the two linear best fits used to find the CMC (see more details in the figure caption). The result of the measurement is that the CMC is 80  $\pm 5 \mu l / l$.

The concentrations  are expressed in $\mu l / l$, with the meaning of volume of the surfactant, as it is found in the vendor sealed container, per liter of water. We don't know the concentration of the surfactant, expressed in mass per volume, as it is in the vendor's container, however in our measurement of the CMC we have expressed the concentrations in the same way (volume of vendor's concentration per volume of water).
This allows to express the concentrations as a fraction of the CMC, and this will also be used later in \cref{fig:summary_scatter}.

\newpage

\subsection{Methods for image processing}
\label{sec:methDL}

As described in the introduction, the wet area measurement has been performed using imaging methods: a drop of the water-based solution has been deposited on a leaf, using a manual adjustable  pipette, with volume set to 10 $\mu l$ 
 and  its spreading process has been video-recorded and analyzed.

Several samples of a solution of colloidal silver in a fixed concentration, and with different concentrations of silicon-based surfactant have been prepared. Then a drop of each surfactant concentration has been deployed on a clean, dry  cucumber leaf in horizontal position, and the spreading has been recorded.
The images have been obtained with a  microscope Zeiss model AXIO Zoom.V16 (Germany),  equipped with a PlanApo Z 0.5x/0.125 FWD 114 objective operated  at 3.5 (i.e. minimal) optical magnification, and the digital video has been acquired with a camera Zeiss Axiocam 506 color integrated in the optical system, with a frame rate of 30 frames per seconds.

We have developed  software tools, written in Python, to perform the processing of the video recordings.
The first step has been the extraction of the frames from the video files; then the Deep Learning (DL) model adapted from the HED-UNet model has been implemented using the PyTorch DL framework, to process the single frames (still images) from the video, in particular for the segmentation, assigning each pixel to one out of two categories: wet and not-wet.

After the segmentation, the number of pixels of  the wet area of each image have been computed, and the wet area has been computed using the appropriate conversion coefficient, taking into account the optical magnification.

The data of the timestamp of each frame, and the area estimate, have been recorded in a data file, for each value of the surfactant concentration.

The data has been further elaborated, to  study the effect of the surfactant on the spreading of the water-based pesticide solution.

\subsection{Numerical comparative algorithms}
\label{sec:algo-discuss}

As briefly mentioned in the introduction, other options for the segmentation exist, other than the manual segmentation. So in this section they will be discussed, and compared  with the Deep Learning approach.
In a previous publication it has already been reported about other attempts from our group  with some numerical algorithms  for the automatic image segmentation \cite{Grazioso-22}.


The numerical algorithms for segmentation rely on the numerical analysis of the values of  intensity, and/or saturation, luminosity, hue etc. of each pixel, and their comparison.

So, at the beginning it is needed to analyze a certain number of images manually, so to find some common numerical rule that can effectively differentiate the pixels from the wet  and the dry areas of the leaves.

Then, the numerical algorithms can be grouped into two general classes: those that rely on some threshold value and those that rely on some differential computation. 

The algorithms in the first group compare each pixel to an intensity threshold value, in order to determine to which class the pixel belongs (in our case, wet or dry surface).  Regarding this category of algorithms, it can be observed that this approach is not ideal when the images come from a \emph{general purpose camera}. Indeed, most of such cameras have a mechanism that dynamically change the gain of their CCD, depending on the amount of dark and bright areas present in the subject. The result of this feature is that the brightness and intensities of the pixels in the different frames of a video may change continuously, depending on the bigger or smaller amount of  dark and clear areas present in the field of view. Moreover, even on a single image, the threshold intensity of the pixels may not be uniform due to some non-uniformity in the illumination and/or irregularity in the surface. 

The overall result is that the optimal threshold can change substantially from image to image, and from region to region of the same image, making the algorithms based on thresholds not very accurate for this particular application.

Regarding the second class of algorithms, which relies on  differences   between  neighbour pixels, they are much more accurate because they are able to compensate for the difference in brightness, from image to image and from region to region of the same image. However, due to the complexity of the computations required, especially if they consider not only the first neighbours of each pixel but also some longer range, the algorithms from this category are computationally intensive, need powerful computers and have in general lower efficiency. Moreover, the algorithms based on differential computation usually need some external information about the direction along which to compute the differentiation, and this can be sometimes problematic, and it puts an extra burden on the side of the user and makes this category of algorithms less practical.

\begin{figure}
	\centering
	\includegraphics[width=0.4\linewidth]{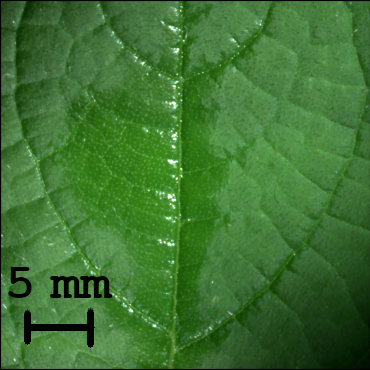}
	\includegraphics[width=0.4\linewidth]{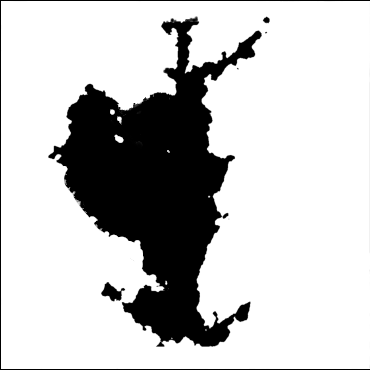}
\caption{Example of a single frame from a video recording, on the left, and the correspondent output of the Deep Learning model segmentation, on the right. In this example, in the center, due to the geometry of leaf, a series of several bright reflection spots are present, aligned in a vertical row, along big part of the image. On the right we can see that the Deep Learning model has correctly identified the reflection bright spots as belonging to the wet area.}
	\label{fig:correct}
\end{figure}

Moreover, if we analyze in particular the images used in this work, we can notice that the wet area has frequently some bright spots which reflect the light of the illumination source, and these spots have a very bright and clear color, that is usually wrongly considered as part of the dry area. This situation can be seen e.g. in \cref{fig:correct}. More in general, we have noticed several mistakes due to the uneven surface of the leaves: the light source can never illuminate evenly the surface, creating brighter and darker spots, that confuse the numerical algorithms.

A possible workaround for this type of mistakes  would be to consider all the pixel topologically internal to a wet area, to  belong to the wet area. In this way, the bright spots due to the reflection would be correctly included in the wet area, although their brightness and color values would assign them  to the dry area.
However, this approach is not viable, because in several cases the spreading of the water is such to leave dry spots completely surrounded by the wet area. Therefore, with this ``topological'' rule, the numerical algorithm would wrongly consider the dry ``islands'' as wet. 

Finally, the presence of dark background with intensity similar to the wet area can easily confuse the numerical algorithm.

In  conclusion, these limits in accuracy and efficiency have motivated the authors to apply  Deep Learning (DL) models to the image segmentation \cite{Liakos2018,GarciaGarcia2018}.

In particular, we have chosen the Deep Learning model called HED-UNet \cite{Heidler2022}, which combines and integrates two different models, one for Semantic Segmentation, and one for Edge Detection.

\begin{figure}
	\centering
	\includegraphics[width=0.4\linewidth]{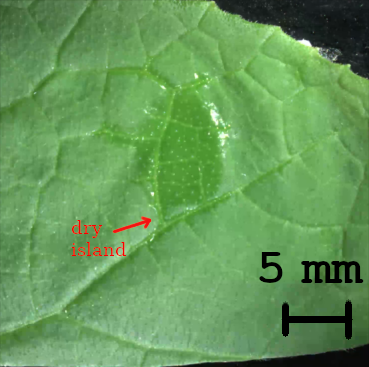} \includegraphics[width=0.4\linewidth]{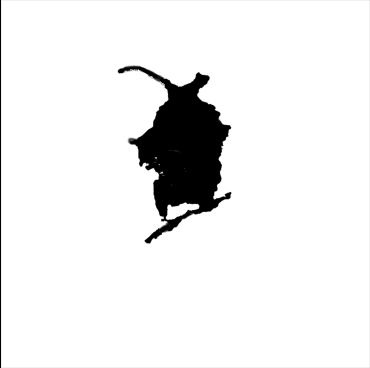}
	\caption{Another example of a single frame from one of the video footage, and the correspondent mask recognized by the Deep Learning model. In this example, we can see how the Deep Learning model has been able to recognize as ``dry'' a spot of dry leave surface, all encircled by wet surface, at its bottom (pointed by the red arrow). In this image we can also see how the DL model can correctly neglect the black background present in the image.}
	\label{fig:DL_leave}
\end{figure}

In  \cref{fig:correct}  we show an example of an image with some bright reflection spots, which have been correctly identified as wet by the DL model, while in  \cref{fig:DL_leave} we show another example of image, with the DL prediction, with an isolated dry ``island'', and the presence of dark background, both correctly identified by the DL model.

\section{Deep Learning}

\subsection{HED-UNet model}

The present work is based the HED-UNet DL model \cite{Heidler2022}.

To apply DL to the segmentation needed for the present work we have decided to apply to this task a DL model already existing, but which was designed for automatic processing (segmentation) of  satellite imagery for geographic and meteorological applications. So, the contribution of the present work in this respect is the idea to apply an existing model to a task completely different from the one for which it has been originally developed.
In this section  some of the main features and characteristics of the model are reported, and  the reader is referred to the original article for a complete description of it.

The strong characteristic of  this DL model  is that it combines the capabilities and the advantages of two different, previous models, the Holistically-Nested edge Detection (HED) model \cite{Xie2015},  and the  UNet model \cite{Ronneberger2015}. Those two older models are designed for two different tasks, namely the \emph{Edge Detection} (HED) and the \emph{Semantic Segmentation} (UNet). 
Although those are two different and separate tasks, at a second look they have a lot in common.
Indeed, to segment an image it means   in essence to separate different areas of the image, i.e. to group in different classes the pixels of the different areas. It is not difficult to show that this is tightly linked to the task of  finding the pixels on the boundaries between  those said areas.
To highlight the strong link between the two tasks, it is worth to notice that both  have the same type of output: a ``mask'' of the same size of the input image, where each element specifies the class of the corresponding pixel in the image.

In the present work, the classification has been designed using only two classes: the class of pixels representing the water, and the class of pixels representing ``all the rest'', dry areas of leaves, and possibly background, if present.

\begin{figure}
	\centering
	\includegraphics[width=0.95\linewidth]{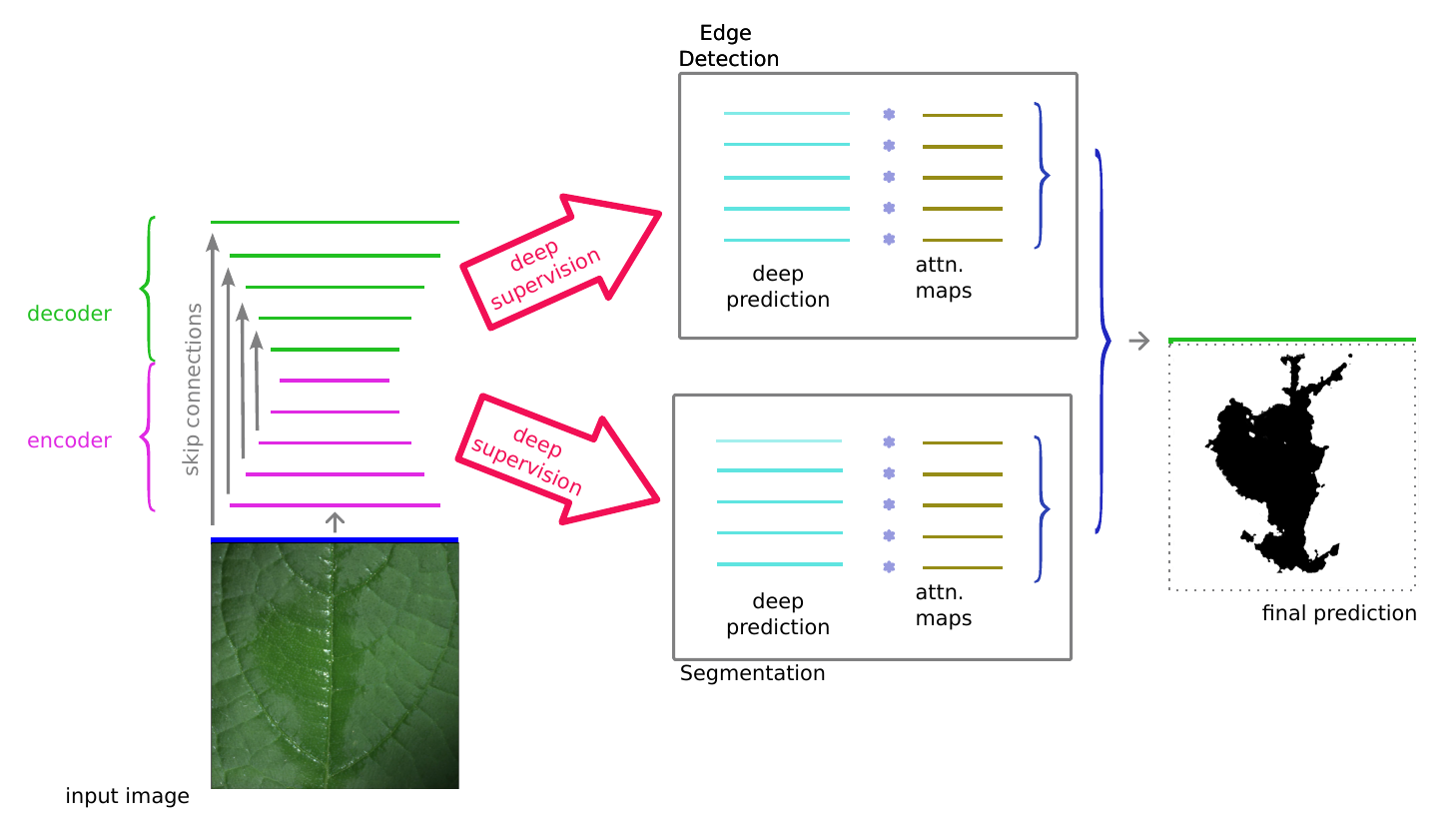}
	\caption{Schematic of the structure of the HED-UNet model (based on the schemes present in \cite{Heidler2022}. In the lower left corner  a depiction of a possible input image is shown. Only this image, and the final prediction mask, are shown in this schematic in their extension:  all the other images, feature masks, and intermediate predictions are represented just as lines, as if the images were seen from an edge. The input image is fed into an encoder-decoder structure, which consists of in subsequent hidden layers that first reduce the original size of the image, and then re expand it, back to the original size (n.b. the encoder-decoder used here consist of 6 hidden layers of lower resolution, the number of layers depicted is smaller for graphic clarity). The encoder-decoder structure makes it possible to exploit both the localized information gathered in the high resolution layers, and the global, nonlocal information obtained from the low resolution layers. The skip connections are parts of the network that connect the encoder layers and the decoder layers. One of the most relevant features of the HED-UNet model is the ``deep supervision''; it consists of part of the network that extract information from the lower resolution layers, independently, and generate separate predictions. This mechanism is used for both the edge detection and the segmentation parts of the model.
		Another relevant feature of the model is the ``attention mechanism''. This consists in special 'attention maps', i.e. weight coefficients, that are multiplied, pixel per pixel, against the intermediate predictions. In this way  different areas  contribute in different ways depending on different degrees of 'importance'. After combining all the intermediate prediction outputs, the final prediction is obtained.}
	\label{fig:model}
\end{figure}

The main components of the HED-UNet model architecture are summarized in  \cref{fig:model}.

The authors of \cite{Heidler2022} have applied this intuition to a practical image-processing problem: find the shore line in satellite pictures of Antarctica landmass, obtaining good results: the ``combined model'' has the two HED and UNet parts in it, which both generate a prediction map which are then combined for a final result.

One of the important parts of the HED-UNet model, already used in  HED, is the so-called \emph{deep supervision}, introduced in \cite{Lee2015}. This technique is proven to improve the result of DL models, and consists in  providing  direct supervision to the intermediate layers, rather than providing supervision only at the output layer.

Moreover, the HED-UNet model (as the HED  and the U-net model) has an ``encoder-decoder'' structure, where a series of downsampling steps (encoder) generate maps at lower and lower resolution, allowing for the aggregation of contextual information, and then a series of upsampling steps generate a series of maps which redistribute this information at a higher resolution, until the original resolution of the input image is restored.
In this scenario, the intermediate maps at lower resolutions are the hidden layers to which the deep supervision is applied, making it a ``multiscale deep supervision'', which is a concept related to the feature pyramid networks discussed in \cite{Lin2017}. \\

\subsubsection{Deep Supervision and Attention mechanism}

In the ``Encoder-Decoder'' module, the input image is reported in different hidden layers with 6 different  resolutions (lower than the input), a structure that is described as a \emph{pyramid}. 
The HED-UNet model has a feature called \emph{Deep Supervision}, which consists of   single \emph{side classifiers} built on each different intermediate resolution layer.
The outputs of each of those side classifiers is merged, to give one final global output. 

There are two  merging modules  (merging heads), one for the Edge Detection part, and one for the Segmentation part of the model. 

Moreover, the model incorporates a variant of the  ``attention mechanism'', originally developed for text processing \cite{Bahdanau2015}, here applied to image processing.

The authors of \cite{Heidler2022} have called this \emph{hierarchical attention merging mechanism}, and it is based on \emph{saliency weight masks} which are computed so to  give different weights to not only to different areas of the same level, but also different weights to different  resolution layers (see also \cite{Wang2018,Niu2021}).
A graphical depiction of the HED-UNet model, including the Deep Supervision and the Hierarchical Attention   mechanism are shown in  \cref{fig:model}.

\subsubsection{Loss function}

When training for edge detection, the model is trained  against two classes, the class of edge pixels and the class of non-edge pixels. However, those two classes are usually highly unbalanced: the number of pixels in the first class are much less than those in the second class.

So, during the training process, the impact on the overall training process, from one type of pixels and the other, is also very unbalanced, and this leads to a much less effective training.

To alleviate this problem, it is possible to design loss functions that compensate for this unbalance.

Already in \cite{Xie2015}, it is used  a \emph{class-balanced} version of the binary cross entropy loss function. Since  also the application of the present work uses two classes in its segmentation (``water'' and ``not-water'' pixels), in the present work, following \cite{Heidler2022}, the same loss function is used, which will be described later in some detail.

Here it is denoted with: 
\begin{equation}\label{eq:vectors}
	\vec{X}_n = \left\{ x_i^{(n)} \right\}_{i=1}^{\left|\vec{X}_n\right|}  \qquad  \text{and} \qquad   \vec{Y}_n = \left\{ y_i^{(n)} \right\}_{i=1}^{\left|\vec{Y}_n\right|}
\end{equation}
respectively the  vector of all the $\left|\vec{X}_n\right|$ pixels values from the $n$-th image used in the training set, and the vector of $\left|\vec{Y}_n\right|$ values of the corresponding ground truth.
The values for the ground truth ``pixels'' in our case are only two: $\forall i, y_i^{(n)}\in\{0, 1\}$, because we have two classes (``water'' and ``all the rest'').
Moreover, it is denoted with $\vec{W}$ the set of all the weights and other parameters of the whole neural network of the model.

Finally, it is denoted with $\left\{\hat{p}_{i}^{(n)}(x_i, \vec{W}) \right\}_{i=1}^{\left|\vec{X}_n\right|}$ the set of predictions values for the $n$-th input image, i.e. the $\left|\vec{X}_n\right|$ output values of the neural network, where $\left|\vec{X}_n\right|$ is the total number of pixels in the image.
This prediction function $\hat{p}^{(n)}$ is a value $\hat{p} \in [0, 1]$, and can be interpreted as the 'probability' of the correct output being equal to 0, or being equal to 1: $\hat{p}(y_i=0|\vec{X}, \vec{W}) $ is the probability of the correct $i$-th 'output pixel' being 0, given the values of (all) the input pixels $\vec{X}$, and (all) the weights of the network $\vec{W}$ (for simplicity we have omitted the index $n$ identifying the image). Similarly, for $\hat{p}(y_i=1|\vec{X}, \vec{W}) $. In this special case where the output can be only 1 or 0, we have $\hat{p}(y_i=0|\vec{X}, \vec{W}) = [ 1 - \hat{p}(y_i=1 | \vec{X}, \vec{W})]$, therefore we can use only one value, which we will denote briefly as $\hat{p}_i^(n)$.

The training process consists in finding the values for the weights $\vec{W}$ such that the set of all the prediction values $\left\{\hat{p}_i^{(n)}\right\}_i$ are as close as possible to all the ground truth values $\left\{ y_i^{(n)} \right\}_{i}$, for all the pixels in all the images (i.e.$\forall i, \forall n$) \cite{Nielsen2015,Grazioso2022a}.
So, the training process consists in the minimization of the \emph{loss function} that measures quantitatively the ``global distance'' between (all the) sets of predictions and (all the) sets of ground truths. 

One initial option for such ``distance estimating'' loss function, in the case where there are only two possible classes, is the \emph{Binary Cross Entropy} function:
\begin{equation}\label{eq:BCE}
 \mathcal{L}_{bce} =	\sum_n \sum_i \left[ y_i \cdot \log (\hat{p}_i^{(n)}) \right] + \left[(1-y_i) \cdot \log (1-\hat{p}_i^{(n)})\right].
\end{equation}

In the case where the number of pixels in the two classes are highly unbalanced (water pixels much less than non-water pixels), it is possible to compensate, and re-balance it as follows: 
\begin{equation}\label{eq:balancedBCEloss}
	\mathcal{\tilde{L}}_{bbce} = - \left[\beta \cdot \sum_{n, i \in Y_1} \log \left[\hat{p}(y_i=1)\right] \right] - \left[(1-\beta) \cdot \sum_{n, i \in Y_0} \log\left[ \hat{p}(y_i=0) \right] \right]
\end{equation}
where the first sum is computed only on the pixels with a ground truth = 1, and the second sum only on those with a ground truth = 0, and where 
\begin{equation}
	\beta = \frac{\left|Y_0\right|}{\left|Y\right|}
\end{equation}
is the class-balancing weight that allows the automatic re-balance of the two classes, where \( Y = Y_1 \cup Y_0 \); in essence $\beta$ is the ratio between the number of pixels in the first class (water), with respect to the total number of pixels.

The base for the loss function used in this project for the training has been the Binary Cross-Entropy with Logits Loss, from the PyTorch library (\texttt{torch.nn.BCEWithLogitsLoss()}). 

\subsection{Training}

We have performed the training of the model in our facilities.

130 images have been  manually annotated, selecting images as diverse as possible, form different videos of different leaves, so to diversify the training and have better predictive performance of the trained model.
 80\% of those images have been used as training set, and 20\% of them have been used as validation set.

For the manual annotation it has been used an image editing software (GIMP, free/libre and open source software) and a graphic tablet (Artist 24 Pro, XP-Pen, Japan-China). 

For each image in the training set, a superimposing black-and-white mask has been created (ground truth),  indicating the wet leave surface pixels and the pixels representing dry leave surface or background.

Then, some \emph{data augmentation} has been used: each image and corresponding mask have been rotated in 4 possible angles, and mirrored horizontally and vertically, so that for each frame 6 different training images have been generated, for a total of $130 \times 6 = 780$ images and masks.

The training of the model has been performed on a remote server, equipped with one NVIDIA Tesla ``V100s'' graphics card, with 32 Gb of video RAM. The training of 780  images, for 320 epochs has taken around 2 days of  computing.

\begin{figure}
	\centering
	\includegraphics[width=0.7\linewidth]{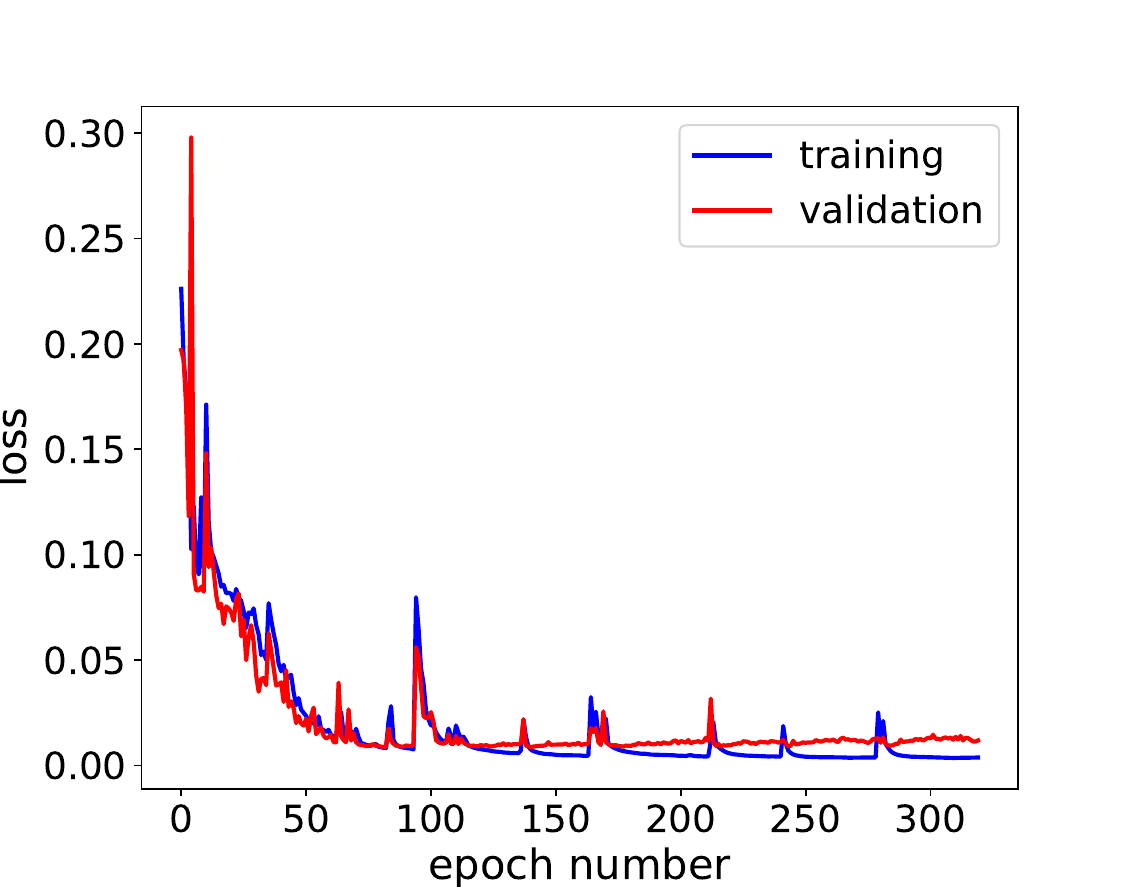}
	\caption{\label{fig:loss} Plots of the training loss and the validation loss as functions of the number of epochs. We can observe a stable loss of around 0.001, computed with the \texttt{torch.nn.BCEWithLogitsLoss()} loss function explicitly reported in equation \eqref{eq:balancedBCEloss}.}
\end{figure}

In  \cref{fig:loss} it is shown the plot of the training loss and the validation loss as functions of the epochs. 	At the end of the training  we have a stable loss of around 0.001.

Once trained, the network is able to perform the segmentation of a single image in around one second.

\subsubsection{Image resizing}

In the encoder-decoder module, the HED-UNet model performs several \emph{resizing} (downsampling/upsampling) on the input image, halving the number of pixels (for both width and height) for each downsampling, and doubling it back for each upsampling. This creates a requirement for  the dimension of the input images: they need to have a number of pixels (both in height and width) that can be divided by two several times with an integer result. 

The root of this requirement is the fact that the number of pixels can only be an integer. Therefore, if a module would halve an odd dimension, it would need to round the result to the next integer; in the up-conversion steps, when the dimensions are doubled and brought back to the initial dimensions, this would result in a mismatch in the number of pixels.

To solve this problem, a pre-processing step is needed, resizing  the images and the ground truths to a suitable dimension before they are fed into the neural network. A square dimension, with a side of $1024$ pixels has been used, where 1024 is a power of 2, so that the division by two always gives an even result.

\newpage

\section{Results and discussion}

\subsection{Wet area evolution}
The main result of this work is the study of the evolution of the wet area on cucumber leaves, and how this is affected by the concentration of the surfactant. Indeed,  surfactants  are used specifically to change the wettability of the pesticide. We have explored the range of concentrations starting from about half the CMC, up to around 11 times the CMC.

\begin{figure}
	\centering
	\includegraphics[width=0.6\linewidth]{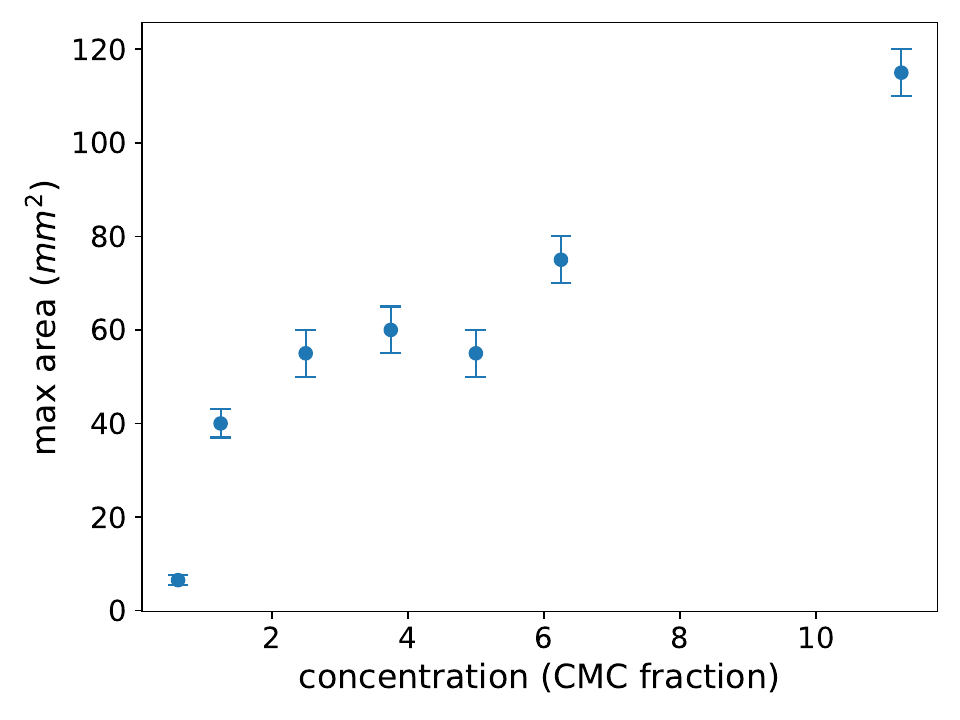}
	\caption{Summary plot of the estimated maximum areas, for each surfactant concentration. We have estimated the maximum extension reached by the wet area, when it reaches a plateau, before it starts to decrease due to evaporation. The error bars represent the range of variation of the area values, that have been estimated within the relatively constant plateau. The concentrations are expressed as fractions of the CMC of the surfactant (that we have directly measured). The values for the concentrations are	 50, 100, 200, 300, 400, 500, and 900 $\mu l / l$, whereas the CMC is 80 $\mu l / l$. So, these concentrations are 0.625,  1.25,   2.5,    3.75,   6.25 and  11.25  expressed as fraction of the CMC.}
	\label{fig:summary_scatter}
\end{figure}

\begin{figure}
	\centering
	\includegraphics[width=0.7\linewidth]{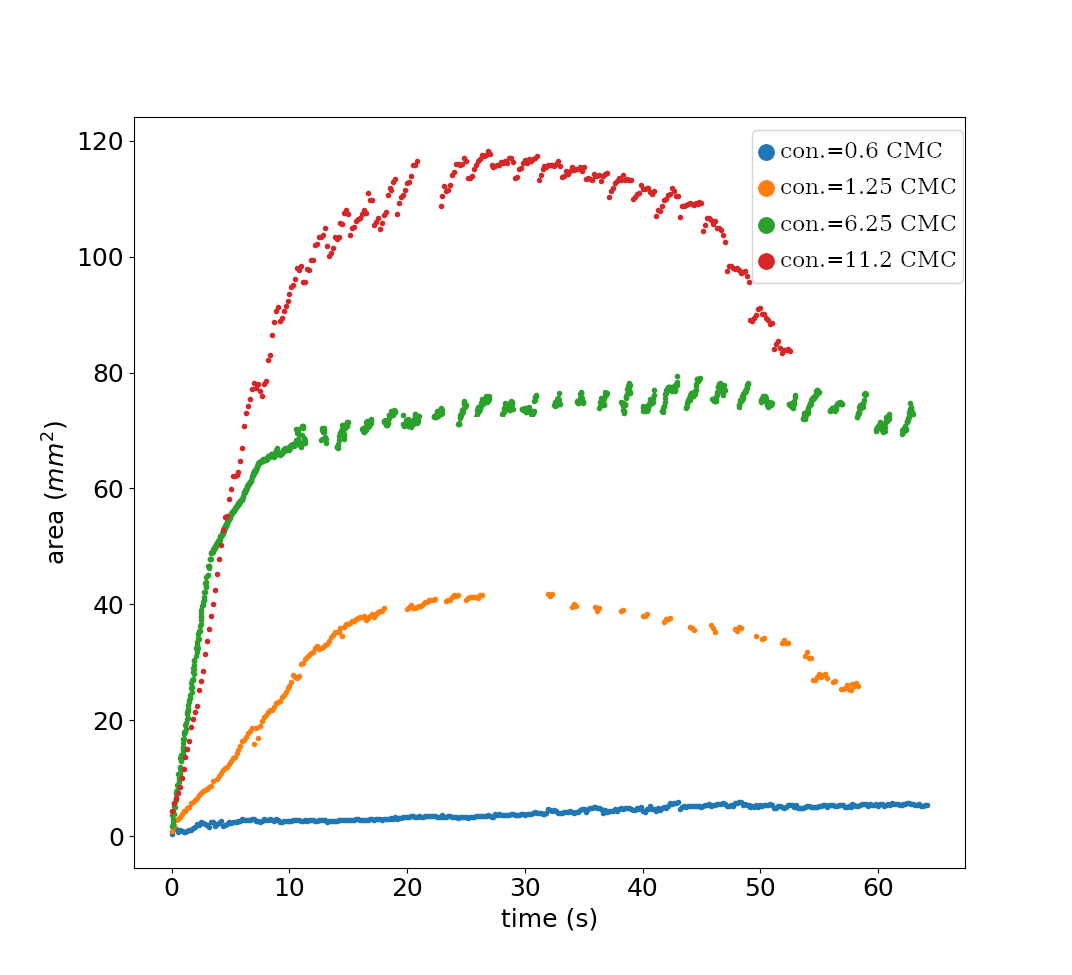}
	\caption{In this plot we have a summary and a comparison of the behaviour of wet area spreading over time, for different concentrations of the surfactant. The concentrations in the legend are expressed in $\mu l / l$. We can observe that at the lowest concentration the solution spreads very little. Moreover, we observe no reduction of the area in the final part. This is due to the fact that the solution with low concentration is so hydrophobic that it layer remains thick above the leaf surface, and therefore it doesn't evaporate much, at least within the time duration of this measurement.}
	\label{fig:summary_plot}
\end{figure}

Qualitatively, we have observed that for all the concentrations except the lowest, the area reaches a maximum expansion value, stays on that value for a while, and then decreases, due to evaporation.
At the lowest concentration the solution spreads very little, and we observe  no plateau and reduction of the area in the final part. This is due to the fact that the solution with low concentration is so hydrophobic that its layer remains thick above the leaf surface, and therefore it doesn't evaporate much, at least within the time duration of our measurement.

In \cref{fig:summary_scatter} we show a plot of the estimated maximum areas, for each surfactant concentration.  We have  expressed the concentrations as fractions of the CMC of the surfactant, which has been measured as  $80 \pm 5 \mu l / l$ (see \cref{sec:CMCmeas} and   \cref{fig:CMC}).

In \cref{fig:summary_plot}  we compare side by side the behaviour of wet area spreading over time, for different concentrations of the surfactant. 
We observe a very small  and almost constant value of the area for the first concentration, if compared with the other curves at different concentrations.

\begin{figure}
	\centering
	\includegraphics[width=0.45\linewidth]{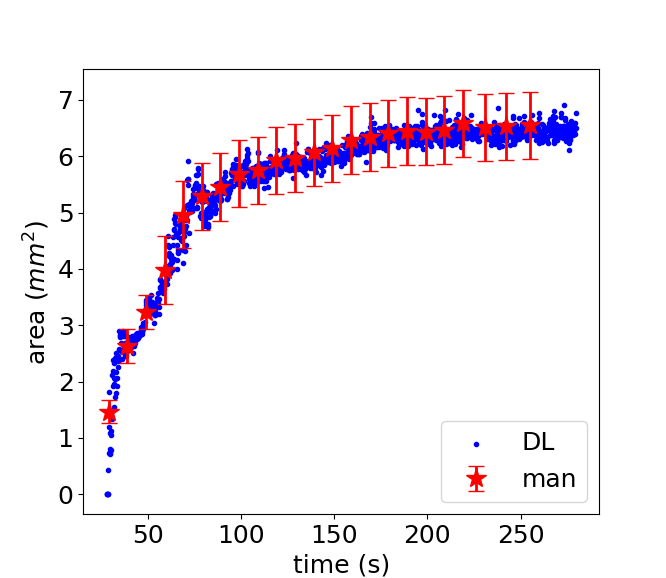}
	\includegraphics[width=0.45\linewidth]{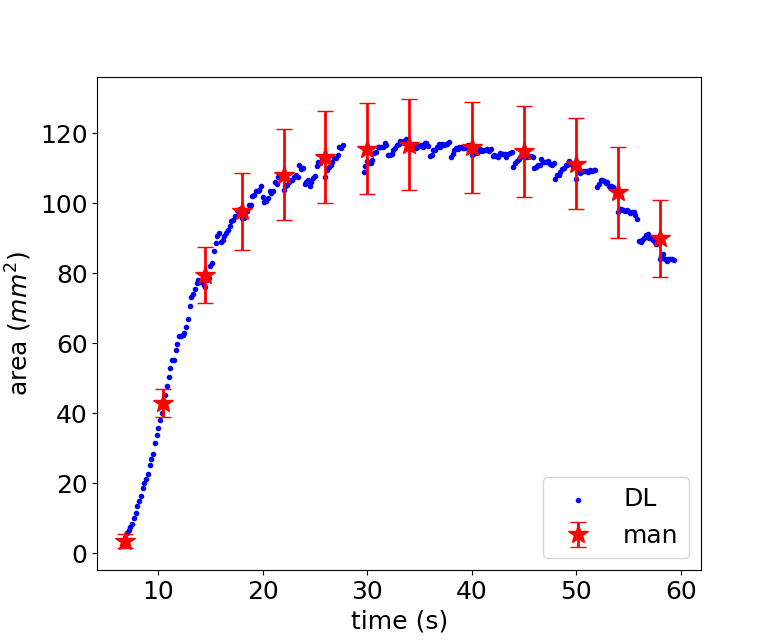}
	\caption{Examples of the progress of wet area over time. On left we have  the data referred to the lowest concentration used in the experiment, 50 $\mu l / l$, wich is equal to 62\% of the CMC. In this case the initial drop doesn't spread much. Therefore we have a rather small area, and we do not observe the decrease of area although in this case the time period observed is much longer, because the water layer remains thick and evaporate less. On the right we have the data referred to the highest concentration used, 900 $\mu l / l$, which is equal to 11.2 times the CMC. Here we can observe a fast spread, a plateau of relatively constant area, at the maximum spread, and the area drop, due to evaporation. The small (blue) dots represent the area measured with the help of the Deep Learning image processing, while the red stars represent the manually measured area.
		The error bars of the manually measured area have been estimated by repetition of the measurement, i.e. manually measuring several times the same image, with different rotations and flipping of the image. The standard deviation of the small sample of measured values has been used to infer the standard deviation of the normal distribution of measurements using the Student's distribution.}
	\label{fig:area_plot}
\end{figure}

In \cref{fig:area_plot} we report  the evolution of wet area over time for just two concentrations, namely the lowest (below CMC) and the highest (around 11 times the CMC). 

In those plots we also compare the area measured with the DL image segmentation, with the area measured with manual image segmentation (manual measurement).
We have  made an estimate of the error affecting the manual measurement, by means of performing several manual segmentation of the same image, with the same frame taken from a spreading video. To make each of those different manual measurements more meaningful for the estimate of the error, each measured image has beer rotated by a different angle, or mirrored with a horizontal or vertical symmetry. We have obtained a small sample of different manual measurements of the same image, we have measured its standard deviation, and then used it to infer the standard deviation of the normal distribution of measurements, using the Student's distribution.

The DL-based measurements appear to be \emph{consistent} with the manual measurements.

\subsection{Automatic image segmentation}

The present work has explored the use of  a DL model for the automatic image segmentation of the wet area on plant leaves, and compared it with the other possible approaches to the same task.

The other approaches considered are the \emph{manual segmentation}, using a raster graphics editing software such as GIMP  and the \emph{algorithmic segmentation}, i.e. the use of numerical algorithms which directly compare the values of the pixels, using either some threshold on some of the pixels values, such as the RGB color components, and/or their hue, saturation and brightness (HSL) values, or some differential rule that computes the gradients of those values while going from a pixel to its neighbors.

\begin{figure}
	\centering
	\includegraphics[width=0.15\linewidth]{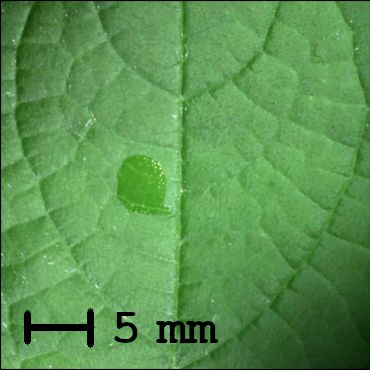}
	\includegraphics[width=0.15\linewidth]{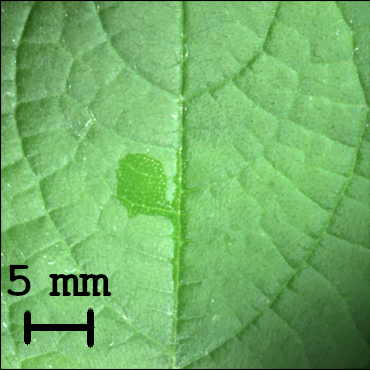}
	\includegraphics[width=0.15\linewidth]{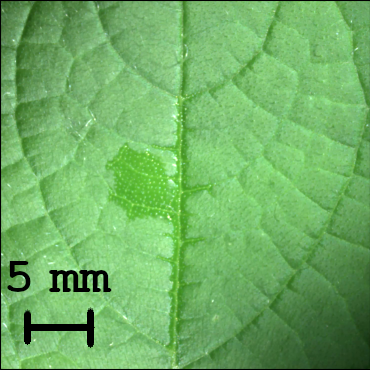}
	\includegraphics[width=0.15\linewidth]{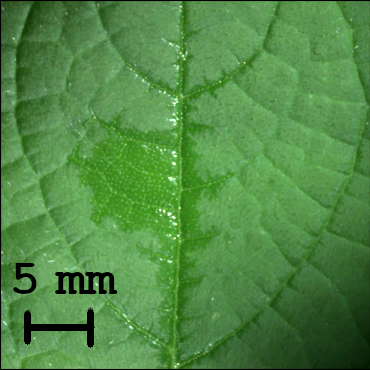}
	\includegraphics[width=0.15\linewidth]{"figs/renamed_c98ae"}
	\includegraphics[width=0.15\linewidth]{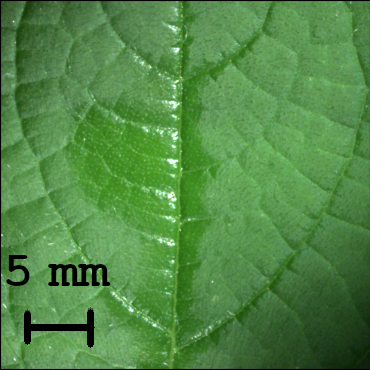}
	\includegraphics[width=0.15\linewidth]{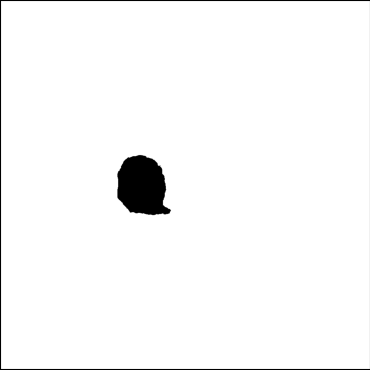}
	\includegraphics[width=0.15\linewidth]{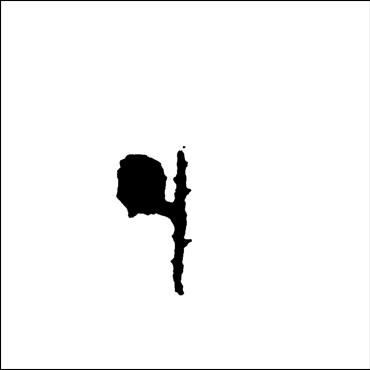}
	\includegraphics[width=0.15\linewidth]{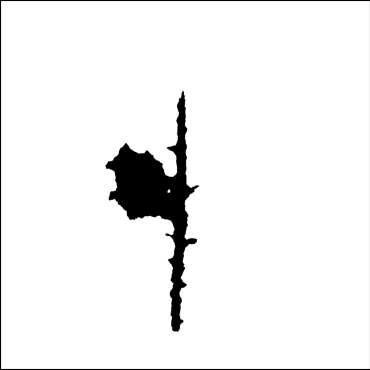}
	\includegraphics[width=0.15\linewidth]{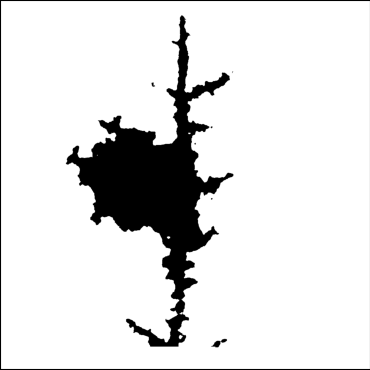}	
	\includegraphics[width=0.15\linewidth]{"figs/renamed_7ca63"}
	\includegraphics[width=0.15\linewidth]{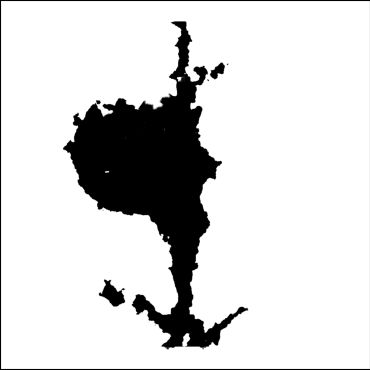}
	\caption{Here we can follow the evolution over time of the wet area on a leaf, at time t =  0,  1,  3,  7,   29 and  58 seconds respectively. From left to right we can observe the initial small and round drop, then the bigger area with more fragmented edges, and how the area of the last two images reduces, after reaching the maximum extension, due to evaporation. For each image in the top row, we show the output of the prediction performed by the Deep Learning model.}
	\label{fig:frames_time}
\end{figure}

\begin{figure}
	\centering
	\includegraphics[width=0.15\linewidth]{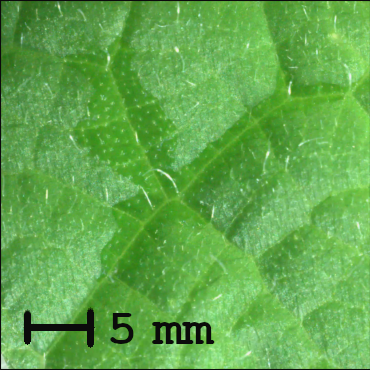}
	\includegraphics[width=0.15\linewidth]{"figs/renamed_c98ae"}
	\includegraphics[width=0.15\linewidth]{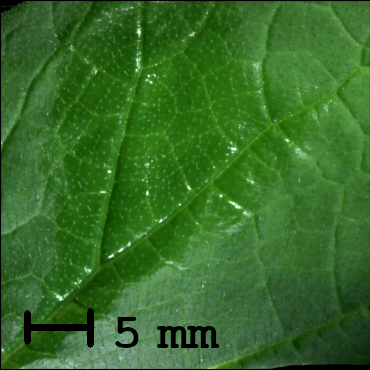}
	\includegraphics[width=0.15\linewidth]{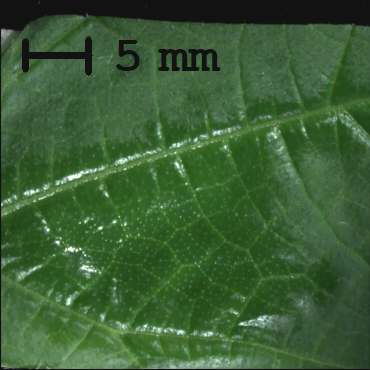}
	\includegraphics[width=0.15\linewidth]{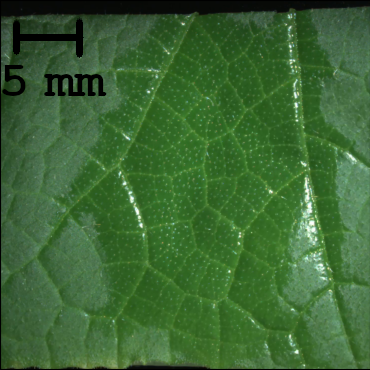}
	\includegraphics[width=0.15\linewidth]{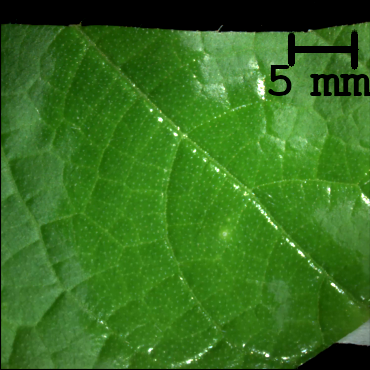}
	\includegraphics[width=0.15\linewidth]{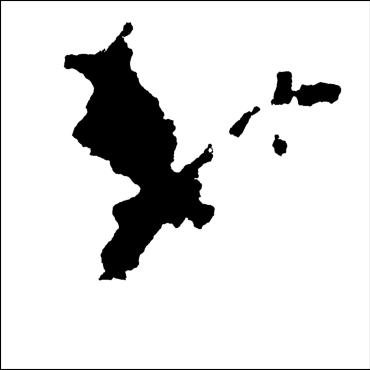}
	\includegraphics[width=0.15\linewidth]{"figs/renamed_7ca63"}
	\includegraphics[width=0.15\linewidth]{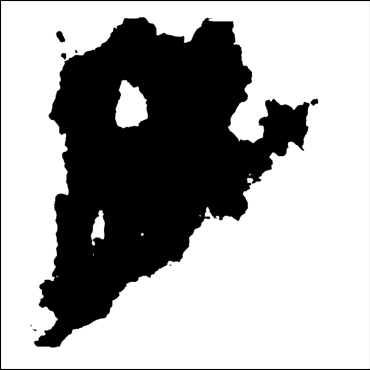}
	\includegraphics[width=0.15\linewidth]{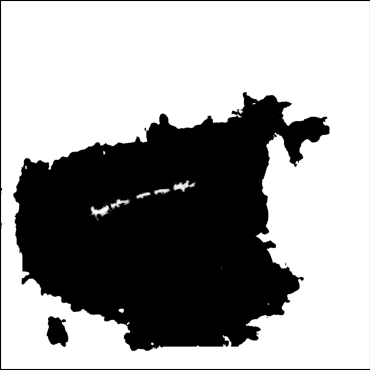}
	\includegraphics[width=0.15\linewidth]{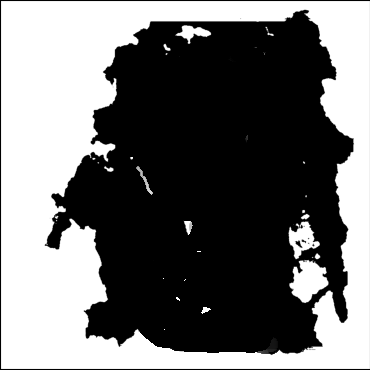}
	\includegraphics[width=0.15\linewidth]{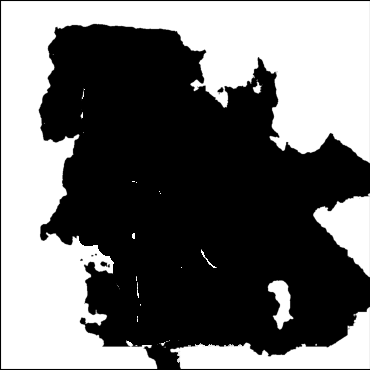}
	
	\caption{Here we can observe the different \emph{maximum area extent} as the concentration increases. In particular, from left to right the pictures refer to the concentrations of 0.625,  1.25,   2.5,    3.75,   6.25 and  11.25 expressed as fraction of the CMC.}
	\label{fig:frames_conc}
\end{figure}

The comparison with the manual segmentation is straightforward. Although the manual process is sometimes more accurate, the time to process a single image, especially those with an extended wet surface with rather intricate and fragmented edges, can take up to 40 minutes, especially if a good level of accuracy needs to me reached. The average processing time can be even longer if we take into account the fact that the process is rather physically and mentally tiring, and it requires some extra time for rests.
In comparison, the processing of a single image with the Deep Learning model takes a time in the order of seconds, with an accuracy that we show in \cref{fig:correct,fig:DL_leave,fig:frames_time,fig:frames_conc}. The accuracy is not as good as the one obtained with the manual segmentation, but it is acceptable, and considering the time needed for the manual segmentation is completely unpractical when the number of images that need to be processed is in the range of thousands.

The comparison with the algorithmic segmentation is less obvious. The comparison with respect to the processing time  gives no significant advantage to neither of the two, although the algorithms based on differential computations tend to be slightly slower, and more accurate, especially on non-uniformly illuminated images, with respect to the algorithms based on a threshold.

But the accuracy of the DL segmentation is significantly higher, in particular with respect to some specific features.
Already in  \cref{sec:methDL} we have reported how the DL segmentation is particularly efficient in distinguishing the ``\emph{islands}'' of dry surface completely surrounded by wet areas, whereas the algorithmic segmentation is usually unable to distinguish them.
Similarly, the DL segmentation is efficient in not interpreting the dark \emph{background} as a wet area, whereas the algorithm usually misinterprets it as wet surface, because it is dark, and it is difficult to process using some threshold values.

We have observed that the  segmentation mistakes of the DL model that we have trained are almost always ``false negative'' mistakes, and almost never ``false positives''. In other words, the pixels representing the dry surface (and possibly the background) are almost always classified as such, whereas the pixels representing the wet surface are sometimes classified as dry.

This means that the wet area measurements based on the DL segmentation are on average underestimated. This has been observed inspecting single images and the corresponding prediction masks (see the examples in \cref{fig:correct,fig:DL_leave,fig:frames_time,fig:frames_conc}), and also confirmed by the comparison between the few  manual measurements  and the DL measurements shown in \cref{fig:area_plot}. Indeed, in this two plots, referred to two concentrations, we can observe how the datapoints representing the manual measurements are positioned at the upper edge of the ribbon of datapoints representing the automatic measurements.

\section{Conclusions}

The contribution of this work is twofold.

First, it reports the measurement of the maximum wet area of pesticide formulation (colloidal silver) over cucumber leaves, as a function of the  concentration of a surfactant. The result of these measurements show that the maximum wet area grows with the  concentration monotonically (see \cref{fig:summary_scatter}), and this is in line with the expected behaviour of the surfactant. More investigation is needed to model the (non linear) function that describes the dependence of the maximum area on the surfactant concentration.

 The second main contribution of this work is the use of a Deep Learning model, developed for a different purpose, and applied here to  efficiently process the high number of images involved, in particular  for the image segmentation.
 Although the model was developed for a different application for satellite images segmentation, its characteristics that combine a sub-model designed for the segmentation and a sub-model designed for the edge detection, make the idea to apply it to this different task of the segmentation of wet area on leaves an effective idea.

So  the second, and most important result of the work is the demonstration of the measurement method, which can be applied to several similar measurements.

A  limit of the present work is the relatively small number of images used for the training of the model. Although one of the specific characteristics of the model chosen is indeed the ability to give good results with relatively small training sets, it may be that this has led to some \emph{overfitting} of the model, which makes it less effective if used with a different dataset.
It should be noted to this regard that the goal of this work is to prove the efficacy of the method, not to develop a software tool readily suitable `as-is' for a large range of different applications, e.g. different leaves or different liquids.

\section*{Acknowledgement}

The research was funded by the Ministry of Science and Higher Education of the Russian Federation - FEWZ-2023-0005.

\newpage
\bibliography{biblio}

\end{document}